\documentclass[journal]{IEEEtran}
\usepackage{amsmath,amssymb,booktabs,graphicx,multirow,url,xcolor}
\usepackage{microtype}
\AtBeginDocument{%
 \setlength{\abovedisplayskip}{2pt plus 1pt minus 1pt}%
 \setlength{\belowdisplayskip}{2pt plus 1pt minus 1pt}%
 \setlength{\abovedisplayshortskip}{2pt plus 1pt minus 1pt}%
 \setlength{\belowdisplayshortskip}{2pt plus 1pt minus 1pt}%
 \setlength{\dbltextfloatsep}{8pt plus 1pt minus 1pt}%
 \setlength{\textfloatsep}{8pt plus 1pt minus 1pt}%
}
\usepackage{dblfloatfix}
\usepackage{placeins}
\usepackage[hidelinks]{hyperref}
\newcommand{\method}{TopoDPO}
\newcommand{\topodpo}{TopoDPO}

\begin{document}
\title{TopoDPO: Dense Preference Alignment for Structure-Sensitive Aerial Referring Segmentation}
\author{Huaiyuan~Peng, Jisheng~Dang, Wencan~Zhang, Jingwen~Zhao,
Hong~Peng,
Wenjun Ma, 
Bin~Hu,~\IEEEmembership{Fellow, IEEE}, Qi~Tian,~\IEEEmembership{Fellow, IEEE}, Tat-Seng~Chua%
\thanks{This work was supported in part by the National Natural Science Foundation of China under Grants 62227807 and U24B20186, in part by the Scientific and Technological Innovation 2030 Project of MOST, China, under Grants 2021ZD0200600 and 2021ZD0200408, and by the Supercomputing Center of Lanzhou University.}%
\thanks{Huaiyuan Peng, Jisheng Dang, Jingwen Zhao, Wenjun Ma, Hong Peng, and Bin Hu are with the School of Information Science and Engineering, Lanzhou University, Lanzhou 730000, China.}%
\thanks{Wencan Zhang and Tat-Seng Chua are with the School of Computing, National University of Singapore, Singapore 119077.}%
\thanks{Qi Tian is with Huawei, Shenzhen 518129, China. Corresponding author: Bin Hu (e-mail: bh@bit.edu.cn).}}
\maketitle

\begin{abstract}
Preference optimization for multimodal segmentation faces a granularity mismatch: dense masks contain spatially heterogeneous preference signals, whereas image-level objectives assign one shared nonlinear weight to all spatial margins. We introduce \method, a topology-aware dense preference alignment framework for structure-sensitive aerial referring segmentation. Topo-Prior mines informative preference pairs using connected-component discrepancy and macro-corner recall, while the dense objective applies the preference penalty to each local margin before spatial aggregation against a frozen supervised reference. On Topo-RS-Hard, a 4590-instance structural stress set, \method{} improves intersection over union (IoU) from 50.91\% to 51.24\% over image-level direct preference optimization (DPO) and reduces Betti-0 error from 1.34 to 0.90; paired image-cluster bootstrap intervals against image-level DPO exclude zero for all five reported metrics. Matched supervised controls and a complete factorial analysis separate the effects of topology-aware pair mining, localized correction, and dense reference-relative preference weighting. An unfiltered stratification over all 39\,244 validation records further shows the clearest structural gains on samples with moderate or severe pre-alignment component-count errors. The code is publicly available at: \url{https://github.com/phy709/TopoDPO}.
\end{abstract}

\section{Introduction}
\IEEEPARstart{R}{eferring} segmentation maps an image and a natural-language query to the pixels of the described object. The task connects language grounding with dense visual prediction and is increasingly studied with multimodal large language models (MLLMs) and promptable segmentation decoders~\cite{lisa,pixellm,sa2va}. In aerial imagery, the same semantic target may be small, elongated, fragmented by occlusion, or embedded in a dense scene. A useful mask must therefore preserve not only region overlap but also the connectivity and geometry of the target.

Remote-sensing foundation models and multimodal systems such as SkySense, GeoChat, and EarthGPT strengthen semantic understanding of overhead imagery~\cite{skysense,geochat,earthgpt}, but they do not specify how a language-conditioned mask should preserve a narrow connection, a hole, or a polygonal corner. Our problem begins at this interface between semantic grounding and structurally faithful dense decoding.

This requirement exposes a weakness of conventional evaluation and alignment. Intersection over Union (IoU) and Dice/F1 aggregate pixels by area. Removing a narrow bridge from a road or adding a small disconnected island may barely change these metrics, although the predicted topology changes. A related issue arises in image-level direct preference optimization (DPO), which averages the spatial log-likelihood margins and then applies one nonlinear preference penalty~\cite{dpo}. Consequently, every coordinate receives the same weight determined by the global mean margin. Positive margins at easy coordinates can cancel negative margins at locally incorrect coordinates before the sigmoid is evaluated. We call this \emph{global-margin coupling} or \emph{local-margin cancellation}. The claim is about aggregation order, not the removal of spatial normalization: both image-level and dense objectives retain a factor of $1/(HW)$.

\begin{figure*}[!t]
\centering
\includegraphics[width=\textwidth]{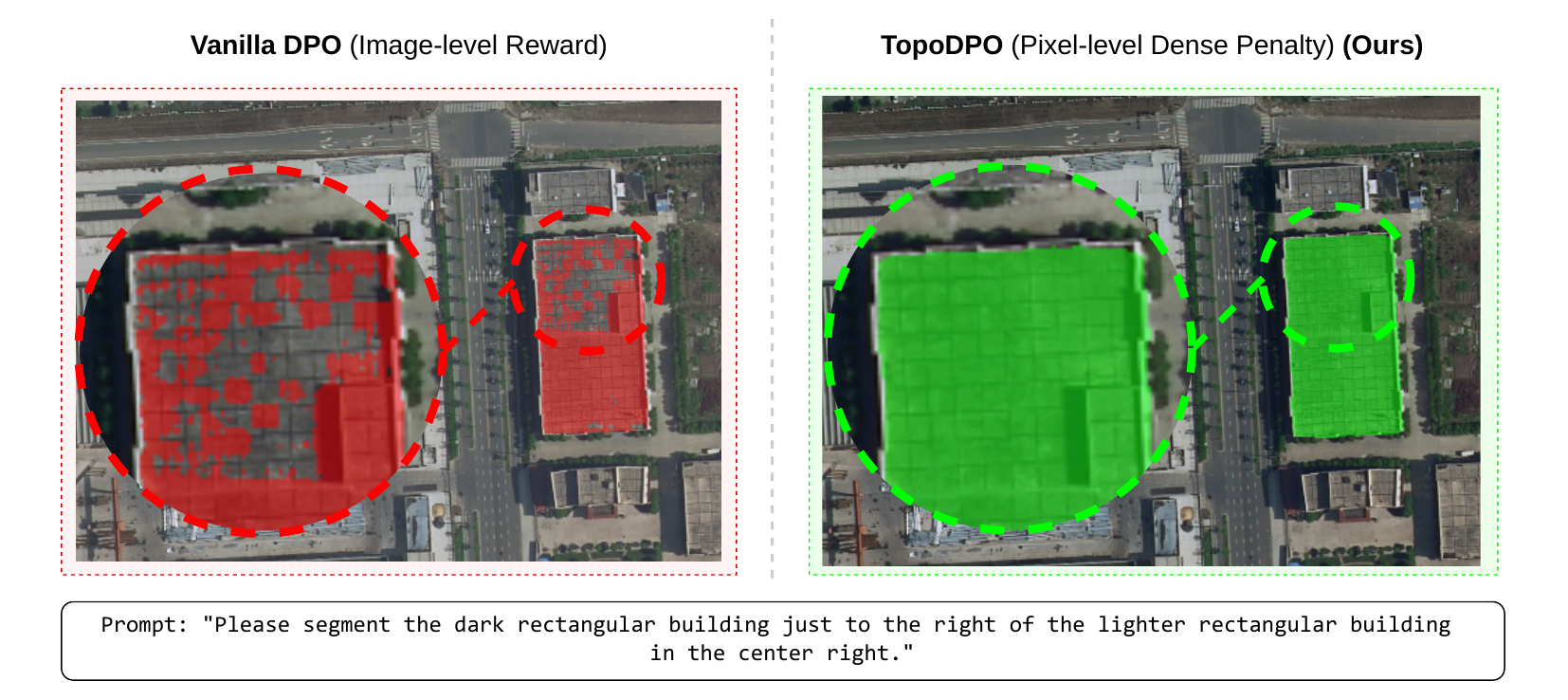}
\caption{Image-level DPO compresses a spatial preference signal into one scalar, whereas \topodpo{} applies the nonlinear preference penalty to the spatial margin before aggregation. The masks illustrate representative localized structural errors addressed by the proposed dense alignment objective.}
\label{fig:teaser}
\end{figure*}

As illustrated in Fig.~\ref{fig:teaser}, image-level DPO aggregates the spatial preference margins before applying the nonlinear penalty, whereas \method{} preserves local margins through nonlinear weighting before spatial aggregation.
We propose \method{} to target these sparse structural errors. The first component, Topo-Prior, evaluates predictions offline using the absolute Betti-0 discrepancy and macro-corner recall, and uses these signals to mine chosen and rejected mask pairs. The second component, \topodpo, preserves the two-dimensional preference-margin map and applies the DPO nonlinearity at each spatial position before averaging. The design does not require differentiating through connected-component or corner extraction during training. It can therefore be integrated with a large language backbone and a segmentation decoder using the same preference pairs as image-level DPO.

We study this design with Sa2VA-InternVL3-8B on Aerial-D and RISORS~\cite{aerial_d,risors}. Topo-RS-Hard is a fixed 4590-instance structural stress set drawn from a 39\,244-instance public validation pool, complemented by an unfiltered analysis over the full pool. A same-protocol clDice baseline, paired image-cluster resampling, matched full-mask supervision, and a gradient-scale-matched disagreement-masked control isolate the effects of topology-aware curation and dense preference weighting~\cite{cldice}.

The contributions are threefold:
\begin{itemize}
\item We identify global-margin coupling in mask preference learning and introduce a spatially dense DPO objective that prevents local margins from canceling before nonlinear weighting.
\item We develop an offline topology-aware preference mining procedure based on connected components and macro corners, avoiding online differentiation through discrete structural operators.
\item We establish the effect with a focused structural stress test and full-pool analysis, topology-loss and matched supervised controls, a factorial ablation of pair selection and objective granularity, and paired image-cluster bootstrap intervals.
\end{itemize}

\FloatBarrier
\section{Related Work}
\subsection{Language-Guided Dense Prediction}
Promptable segmentation models provide pixel-level decoding~\cite{sam,sam2}. LISA and PixelLM connect language-model hidden states to segmentation decoders for reasoning segmentation, while Sa2VA supports dense grounded understanding of images and videos~\cite{lisa,pixellm,sa2va}. GeoChat and EarthGPT extend grounded multimodal understanding to overhead imagery~\cite{geochat,earthgpt}. Our contribution is orthogonal to backbone design and focuses on alignment after supervised fine-tuning (SFT) for structurally faithful dense prediction.

\subsection{Preference Alignment for Multimodal Models}
DPO replaces explicit reward-model training with a classification objective over preferred and rejected outputs~\cite{dpo}. Multimodal variants such as mDPO and task preference optimization adapt preference learning to visual conditioning and differentiable vision heads~\cite{mdpo2024,tpo2025}. POPEN further introduces preference optimization and preference-based ensembling for large vision-language model (LVLM) reasoning segmentation~\cite{popen2025}. These methods establish the relevance of preference feedback to dense multimodal tasks, but they do not target component-level defects in aerial masks.

Recent segmentation-specific studies are the closest comparisons. RN-DPO normalizes mask preferences by their disagreement-region size to stabilize medical segmentation under noisy judges~\cite{rndpo2026}. Active Preference Alignment selects high-disagreement samples and learns a quality judge for Mars terrain segmentation~\cite{activepref2026}. TopoDPO instead uses explicit structural diagnostics to mine pairs and applies the DPO nonlinearity separately to each spatial margin. It retains full-image normalization, so it is complementary to disagreement-region normalization rather than an earlier name for the same operation.
Mask preferences can differ at sparse structural pixels even when most coordinates agree. Unlike token-level methods that align one-dimensional sequences, \topodpo{} retains two-dimensional spatial margins and the same pairwise supervision, avoiding a shared nonlinear weight across easy and difficult pixels.

\subsection{Topology-Aware Segmentation}
Persistent-homology and skeleton-based objectives complement region-overlap losses~\cite{betti_matching,cldice}. Topology-preserving segmentation converts persistence information into an end-to-end loss, while homotopy warping identifies critical pixels using digital topology~\cite{hu2019topology,hu2022homotopy}. clDice compares soft skeletons for tubular structures, and the recent SCNP method couples neighboring logits to prioritize poorly classified local neighborhoods~\cite{cldice,scnp2026}. These approaches demonstrate that similar IoU values can conceal meaningfully different structures. TopoDPO differs by using structural operators for offline pair selection rather than differentiating through a topology surrogate at every update.

Topology-sensitive evaluation is equally important. Boundary IoU isolates contour fidelity that mask IoU can underweight, and average path length similarity (APLS) evaluates route consistency after road masks are converted to graphs~\cite{boundary_iou,apls_metric}. Such measures are valuable but are not interchangeable: a boundary can be locally accurate while a network is disconnected, and two masks can have the same number of components while differing in route connectivity. Our selected diagnostics intentionally remain simple enough to audit across thousands of language-conditioned predictions. Betti-0 captures component-count failures, while macro-corner recall probes coarse polygonal geometry; ordinary overlap metrics remain necessary to detect semantic extent errors.

A recent method named Topo-R1 trains a vision-language model (VLM) with group relative policy optimization (GRPO) to detect and classify synthetic topological anomalies in existing masks~\cite{xu2026topor1}. In contrast, our task is language-guided mask generation, and our output and optimization target are segmentation pixels rather than anomaly boxes and labels. Conventional topology objectives directly train a segmentation network with a differentiable surrogate. Our approach instead uses discrete topology and geometry measures to mine preference pairs offline, then trains with a dense preference objective. This distinction motivates the clDice baseline in our experiments. We report both area metrics and structural measures because neither family alone describes mask quality.

\section{Method}

\begin{figure*}[!t]
\centering
\includegraphics[width=\textwidth]{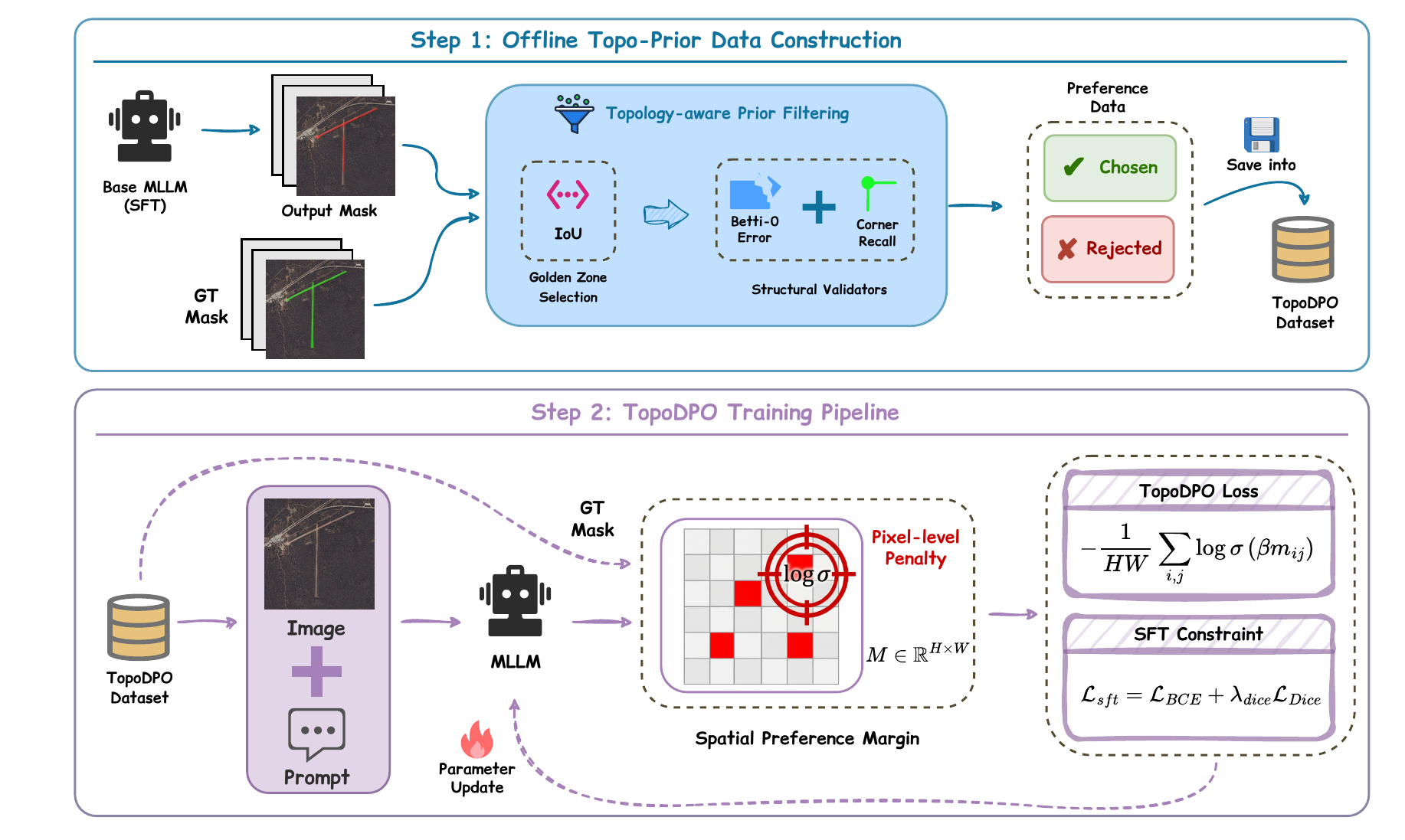}
\caption{Overview of \method. Topo-Prior mines structurally informative mask preferences offline. During training, \topodpo{} computes margins between the policy and reference branches on the mask grid and applies a pixel-wise preference penalty with a supervised segmentation anchor.}
\label{fig:pipeline}
\end{figure*}

\subsection{Problem Formulation}
Given an aerial image $X$ and a referring expression $Q$, a multimodal segmenter $F_\theta$ predicts a binary mask $Y\in\{0,1\}^{H\times W}$. A language model produces the hidden state of a segmentation token; a projection layer converts it to a prompt embedding; and a SAM-2 mask decoder $G$ generates logits
\begin{equation}
M_\theta=G(e_{\mathrm{seg}},\Phi(X)), \qquad \hat{Y}=\sigma(M_\theta).
\end{equation}
For a target mask $Y$, we define the per-pixel log likelihood as the negative binary cross-entropy (BCE)-with-logits loss:
\begin{equation}
\ell_\theta^{ij}(Y)=-\operatorname{BCEWithLogits}(M_\theta^{ij},Y^{ij}).
\end{equation}
A preference example contains a chosen mask $Y_w$ and a rejected mask $Y_l$ for the same $(X,Q)$. We initialize both policy and reference branches from the same supervised checkpoint. The reference policy $\pi_{\mathrm{ref}}$ is a frozen SFT snapshot and is evaluated without gradient updates. Only the policy low-rank adaptation (LoRA) parameters are optimized; the visual encoder, segmentation-token projection, and mask decoder are frozen in both branches. This construction keeps the reference distribution fixed throughout alignment and makes the likelihood ratio consistent with the standard DPO interpretation.

\subsection{Image-Level Preference Objective}
Image-level mask DPO first averages the likelihood map,
\begin{equation}
\ell_\theta(Y)=\frac{1}{HW}\sum_{i,j}\ell_\theta^{ij}(Y),
\end{equation}
and minimizes
\begin{equation}
\mathcal{L}_{\mathrm{img}}=-\log\sigma\!\left(\beta[\Delta\ell_\theta-\Delta\ell_{\mathrm{ref}}]\right),
\label{eq:image_dpo}
\end{equation}
where $\Delta\ell_\theta=\ell_\theta(Y_w)-\ell_\theta(Y_l)$. Equation~\eqref{eq:image_dpo} mixes coordinates before the nonlinear penalty. Its central limitation is therefore not that averaging alone forces sigmoid saturation. Rather, the global mean assigns one common nonlinear weight to all coordinates and permits oppositely signed local margins to cancel before that weight is computed.

\subsection{Topology-Aware Preference Mining}
Topo-Prior constructs mask preferences from supervised-model predictions without additional human ranking. It targets semantically plausible predictions whose remaining errors are predominantly structural. Given a ground-truth mask $Y$ and an SFT prediction $\hat{Y}$, the pipeline first computes IoU and retains the candidate when
\begin{equation}
0.40\leq\operatorname{IoU}(Y,\hat{Y})\leq0.95.
\label{eq:golden_zone}
\end{equation}
Predictions below 0.40 usually contain gross localization or category errors, for which topology refinement is not the primary problem. Predictions above 0.95 contain little disagreement and may be dominated by boundary annotation noise. We call \eqref{eq:golden_zone} the \emph{IoU golden zone}. It controls the semantic difficulty of the preference pairs; the structural tests below determine whether a retained candidate is informative.

\subsubsection{Connectivity Oracle}
Motivated by the use of Betti numbers in topology-preserving segmentation~\cite{edelsbrunner2000topological,hu2019topology}, we measure connectivity by the absolute connected-component discrepancy
\begin{equation}
d_{\beta_0}(Y,\hat{Y})=|\beta_0(Y)-\beta_0(\hat{Y})|,
\label{eq:betti0}
\end{equation}
using 8-connectivity~\cite{rosenfeld1966sequential}. A nonzero value detects both fragmentation, which increases the number of components, and erroneous fusion, which can decrease it. We exclude empty targets and annotations with at least ten connected components so that the component-count signal remains focused on coherent referred entities. Betti-0 therefore provides a compact and auditable connectivity diagnostic for large-scale preference mining.

\subsubsection{Geometric Oracle}
Connectivity alone cannot detect a rectangular building that remains connected but loses its corners. We therefore extract external contours and simplify them with the Douglas Peucker algorithm~\cite{douglas1973algorithms}. The simplification tolerance is 1\% of contour arc length. Let $C_Y$ and $C_{\hat{Y}}$ be the resulting ground-truth and predicted corner sets. Macro-corner recall is
\begin{equation}
R_{\mathrm{corner}}=\frac{1}{|C_Y|}\sum_{c\in C_Y}\mathbb{I}\!\left[\min_{c'\in C_{\hat{Y}}}\|c-c'\|_2\leq\tau\right],
\label{eq:corner}
\end{equation}
where $\tau=15$ pixels at the fixed $256\times256$ mask resolution. A candidate triggers the geometric rule when $R_{\mathrm{corner}}<0.8$. The fixed evaluation resolution and tolerance provide a consistent criterion across all pair construction and evaluation runs while accommodating contour discretization and annotation variation.

The ground truth serves as the chosen mask $Y_w$, and the SFT prediction serves as the rejected mask $Y_l$, when the candidate satisfies \eqref{eq:golden_zone} and triggers either \eqref{eq:betti0} or the corner rule. This procedure yields 8762 preference pairs: 4053 from Aerial-D and 4709 from RISORS, covering 7561 unique images. The recorded selection triggers include 7270 topology-broken pairs and 1492 corner-smoothed pairs. Fig.~\ref{fig:pipeline} summarizes the offline construction and subsequent optimization.

This construction separates diagnostic computation from gradient computation. Connected-component counting and polygon simplification are performed once during data preparation, so their discreteness does not enter back-propagation. The resulting pair retains ordinary mask supervision: the chosen mask is the annotation and the rejected mask is a model prediction for the same image and query pair. This separation makes Topo-Prior modular: structural diagnostics can be changed or extended during pair construction without redesigning the segmentation decoder or the dense preference objective.

The training and evaluation pools are disjoint. Training preferences are mined exclusively from the training splits. Topo-RS-Hard is constructed only for diagnosis from the public validation records of Aerial-D and RISORS. Its non-compactness filter is
\begin{equation}
\mathcal{C}(Y)=\frac{\operatorname{Perimeter}(Y)^2}{\operatorname{Area}(Y)}\geq 40.
\end{equation}
Combining this filter with the structural oracles yields 4590 instances from 39\,244 validation records. Topo-RS-Hard concentrates evaluation on structurally challenging cases, while the complete 39\,244-record analysis in Section~\ref{sec:hard_results} provides the complementary unfiltered view.

The two structural oracles are designed to capture complementary failure modes rather than provide a complete characterization of mask topology. The Betti-0 discrepancy detects changes in the number of connected components, making it sensitive to fragmentation and erroneous fusion. However, it cannot distinguish predictions with identical component counts but different spatial connectivity. Macro-corner recall provides an additional geometric criterion for identifying deviations in coarse polygonal structures, particularly when the predicted mask remains connected. The logical OR between these criteria allows a preference pair to be retained when either type of structural defect is present, without requiring the two errors to occur simultaneously.

The IoU golden zone serves a different purpose. It restricts preference construction to predictions with sufficient semantic overlap while retaining nontrivial mask discrepancies. Consequently, the structural criteria operate on predictions that have already achieved a moderate degree of target localization. The selected thresholds define a fixed mining protocol rather than universally optimal criteria, and their suitability may vary with mask resolution, annotation granularity, and object morphology. Similarly, the exclusion of highly fragmented ground-truth masks restricts the current component-count diagnostic to targets for which this criterion provides a more interpretable structural signal.

Importantly, these criteria determine which prediction errors enter preference optimization, not how individual pixels are weighted during training. The resulting pairs contain ground-truth masks as chosen responses and SFT predictions as rejected responses, while the subsequent dense objective operates on their reference-relative likelihood margins. This separation distinguishes the structural prior used for sample selection from the spatial preference-weighting mechanism and permits the two components to be examined independently through the factorial ablation.

\subsection{Pixel-Level Dense Preference Alignment}
For each coordinate, define the preference margin between the policy and reference branches
\begin{equation}
r^{ij}=\beta[(\ell_\theta^{ij}(Y_w)-\ell_\theta^{ij}(Y_l))-(\ell_{\mathrm{ref}}^{ij}(Y_w)-\ell_{\mathrm{ref}}^{ij}(Y_l))].
\end{equation}
\topodpo{} moves the spatial aggregation outside the nonlinear penalty:
\begin{equation}
\mathcal{L}_{\mathrm{dense}}=-\frac{1}{HW}\sum_{i,j}\log\sigma(r^{ij}).
\label{eq:dense_dpo}
\end{equation}
This ordering retains coordinate-specific margins through the preference nonlinearity. We average over all spatial positions, preserving the same full-image $1/(HW)$ normalization used by image-level DPO. Pixels on which the chosen and rejected masks agree remain in the normalization but contribute zero preference gradient, enabling a controlled comparison that changes only the order of nonlinear weighting and spatial aggregation.

The distinction from image-level DPO is the location of the nonlinear operation. Image-level DPO applies one penalty to the mean margin; two masks with the same mean receive the same loss even if one has a localized severe error and the other has weak errors distributed over the image. Equation~\eqref{eq:dense_dpo} evaluates the local margins first, so a negative margin at a structural coordinate is not canceled algebraically by positive margins elsewhere before the sigmoid is evaluated. Topo-Prior identifies preference pairs exhibiting structural defects, while the dense objective applies reference-relative local correction without differentiating through the offline structural diagnostics. A disagreement-masked BCE control in Section~\ref{sec:hard_results} evaluates conventional supervised correction on the same disagreement support.

We add a supervised anchor to limit drift from the segmentation task:
\begin{equation}
\mathcal{L}_{\mathrm{total}}=\mathcal{L}_{\mathrm{dense}}+\lambda_{\mathrm{sft}}(\mathcal{L}_{\mathrm{BCE}}+\lambda_{\mathrm{dice}}\mathcal{L}_{\mathrm{Dice}}).
\label{eq:total}
\end{equation}
The BCE term retains calibrated per-pixel supervision, while Dice counteracts the imbalance between foreground and background~\cite{milletari2016v}. The coefficient $\lambda_{\mathrm{sft}}$ balances alignment and retention, and $\lambda_{\mathrm{dice}}$ sets the balance inside the supervised anchor. The main configuration uses $\lambda_{\mathrm{sft}}=0.1$ and $\lambda_{\mathrm{dice}}=0.5$. The ablation in Section~\ref{sec:ablation} examines the former.

\subsection{Gradient Analysis}
We next compare how the two objectives propagate a local margin. Let
\begin{equation}
m^{ij}=(\ell_\theta^{ij}(Y_w)-\ell_\theta^{ij}(Y_l))-(\ell_{\mathrm{ref}}^{ij}(Y_w)-\ell_{\mathrm{ref}}^{ij}(Y_l)),
\end{equation}
so that $r^{ij}=\beta m^{ij}$, and let $\bar{m}=\frac{1}{HW}\sum_{i,j}m^{ij}$. For a coordinate $(u,v)$, differentiating the image-level objective gives
\begin{equation}
\frac{\partial\mathcal{L}_{\mathrm{img}}}{\partial m^{uv}}=-\frac{\beta}{HW}\sigma(-\beta\bar{m}).
\label{eq:grad_image}
\end{equation}
Every coordinate is multiplied by the same saturation factor, which depends on the global mean. If most contributions make $\bar{m}$ strongly positive, $\sigma(-\beta\bar{m})$ approaches zero. A locally negative structural margin then receives a small derivative because the objective has already compressed the map.

For the dense objective, the corresponding derivative is
\begin{equation}
\frac{\partial\mathcal{L}_{\mathrm{dense}}}{\partial m^{uv}}=-\frac{\beta}{HW}\sigma(-\beta m^{uv}).
\label{eq:grad_dense}
\end{equation}
The saturation factor now depends on the local margin. When $m^{uv}<0$, its magnitude approaches $\beta/HW$ as the error becomes more severe, even if other coordinates have positive margins. Conversely, an already preferred coordinate with a large positive margin receives little pressure. Equations~\eqref{eq:grad_image} and~\eqref{eq:grad_dense} explain why changing the aggregation order can concentrate optimization on local disagreements.

The same distinction can be stated directly at the loss level. Let $f(x)=-\log\sigma(x)$ and $r^{ij}=\beta m^{ij}$. Then
\begin{equation}
\mathcal{L}_{\mathrm{img}}=f\!\left(\frac{1}{HW}\sum_{i,j}r^{ij}\right),\qquad
\mathcal{L}_{\mathrm{dense}}=\frac{1}{HW}\sum_{i,j}f(r^{ij}).
\end{equation}
Because $f$ is strictly convex, Jensen's inequality gives
\begin{equation}
\mathcal{L}_{\mathrm{dense}}\geq \mathcal{L}_{\mathrm{img}},
\end{equation}
with equality only when the local margins are identical. The dense objective is therefore sensitive not only to the mean preference margin but also to its spatial heterogeneity: a few strongly negative local margins cannot be compensated by large positive margins elsewhere before the nonlinear penalty is applied. This provides a complementary view of the local-margin-cancellation mechanism.

The analysis characterizes the loss-level weighting induced by the two objectives. Both derivatives retain the $1/HW$ factor, and unchanged pixels remain in that denominator. The same full-image normalization is retained in the disagreement-masked BCE control, so the comparison isolates the correction rule on identical disagreement support without introducing sample reweighting by disagreement area. The experiments in Section~\ref{sec:experiments} then evaluate how this local weighting mechanism translates into structural correction at the network level.

\subsection{Relation to Disagreement-Masked BCE}
The chosen mask in our preference pairs is the ground truth. This permits an exact comparison with supervised correction on the disagreement pixels. Let $D^{ij}=\mathbb{I}[Y_w^{ij}\neq Y_l^{ij}]$ and $s^{ij}=2Y_w^{ij}-1$. For Bernoulli mask likelihoods, the local preference margin reduces to
\begin{equation}
r^{ij}=\beta D^{ij}s^{ij}(M_\theta^{ij}-M_{\mathrm{ref}}^{ij}).
\label{eq:binary_margin}
\end{equation}
Thus, for $D^{ij}=1$, dense DPO applies a logistic loss to the \emph{reference-centered} logit $M_\theta^{ij}-M_{\mathrm{ref}}^{ij}$, whereas disagreement-masked BCE applies the same loss family directly to $M_\theta^{ij}$. For $D^{ij}=0$, the dense preference loss is the constant $\log 2$ and has zero gradient. More explicitly, with $t^{ij}=s^{ij}(M_\theta^{ij}-M_{\mathrm{ref}}^{ij})$ and $u^{ij}=s^{ij}M_\theta^{ij}$, their correction gradients are
\begin{equation}
\begin{aligned}
\frac{\partial\mathcal{L}_{\mathrm{dense}}}{\partial M_\theta^{ij}}
&=-\frac{\beta D^{ij}s^{ij}}{HW}\sigma(-\beta t^{ij}),\\
\frac{\partial(\alpha_{\mathrm{mask}}\mathcal{L}_{\mathrm{mask}})}{\partial M_\theta^{ij}}
&=-\frac{\alpha_{\mathrm{mask}}D^{ij}s^{ij}}{HW}\sigma(-u^{ij}).
\end{aligned}
\label{eq:dense_vs_mask_grad}
\end{equation}
Consequently, the dense correction weight depends on the improvement relative to the frozen SFT prediction, while the masked-BCE weight depends on the current prediction alone. At the shared initialization $M_\theta=M_{\mathrm{ref}}$, dense DPO assigns magnitude $\beta/(2HW)$ to every disagreement-pixel logit gradient; masked BCE generally assigns different magnitudes according to the SFT confidence. The fixed factor $\alpha_{\mathrm{mask}}$ in \eqref{eq:mask_scale} matches their \emph{mean} initial absolute gradient magnitude, not their pixelwise weights or subsequent training dynamics. Dense DPO therefore changes the reference-dependent weighting of ordinary GT-labeled disagreement pixels; it does not introduce additional pixel labels or an explicit differentiable topology loss.

\FloatBarrier
\section{Experiments}\label{sec:experiments}
\subsection{Setup}
\subsubsection{Datasets and Splits}
We use Aerial-D and RISORS for aerial referring segmentation and EarthVQA for general remote-sensing multimodal supervision~\cite{aerial_d,risors,earthvqa}. The SFT mixture contains 116\,000 training records: 40\,000 Aerial-D records, a 46\,000-record RISORS stream constructed with controlled repetition to balance the segmentation mixture, and 30\,000 EarthVQA records. Thus, 86\,000 records provide referring-segmentation supervision and 30\,000 provide visual-question-answering supervision.

Topo-Prior operates only on predictions for the Aerial-D and RISORS segmentation training records; EarthVQA records are not used for preference-pair construction. This procedure yields the 8762 preference pairs described above. Quantitative segmentation evaluation is performed on the public validation records with ground-truth masks, and no validation record is used for SFT or preference-pair construction. The combined Aerial-D and RISORS validation pool contains 39\,244 records. From this pool, the fixed Topo-RS-Hard diagnostic subset contains 3721 Aerial-D and 869 RISORS instances (4590 total), corresponding to 2581 unique images; several images have multiple referring expressions. The clDice outputs were matched to the original records by the tuple of image basename and prompt, including multiplicity, before aggregation. This exact multiset match prevents an apparent gain caused by dropping failed or duplicate predictions.

We use a controlled post-SFT alignment protocol in which the image-level DPO and TopoDPO variants share the same SFT initialization, backbone, preference data, optimization schedule, and frozen reference. This design isolates the effect of objective granularity within the matched DPO comparison. Additional published-method baselines are evaluated on the same diagnostic records; the comparisons with these baselines are method-level comparisons unless their training conditions are matched. Topo-RS-Hard provides a focused evaluation of structurally difficult masks, while the complete 39\,244-record Aerial-D+RISORS validation pool provides an unfiltered complementary analysis. For the full-pool stratification, records are grouped only by the pre-alignment SFT Betti-0 discrepancy, with no IoU golden-zone or output-dependent filtering from the aligned models.

\subsubsection{Model and Optimization}
The backbone is Sa2VA-InternVL3-8B, combining InternVL3-8B with a SAM-2 mask decoder~\cite{sa2va,internvl3,sam2}. LoRA with rank 16 and scaling factor 32 is applied to language-model projection layers. Training uses one GPU, micro-batch size 1, and gradient accumulation 16, giving an effective batch size of 16. The implementation uses the XTuner training stack~\cite{xtuner2023}. We train with AdamW~\cite{loshchilov2017decoupled} for one epoch using learning rate $2\times10^{-6}$, cosine decay, and 5\% warmup. For \topodpo{}, $\beta=0.1$, $\lambda_{\mathrm{sft}}=0.1$, and $\lambda_{\mathrm{dice}}=0.5$. The SFT checkpoint is frozen as the reference branch; only policy LoRA parameters are updated, while the visual encoder, projection, and mask decoder remain frozen. Images are encoded at $1024\times1024$, and mask logits are evaluated at $256\times256$.

The image-level and pixel-level DPO comparisons use the same backbone, frozen reference, optimizer, number of steps, and preference pairs. We also include RN-DPO~\cite{rndpo2026}, a region-normalized segmentation preference baseline, and SCNP~\cite{scnp2026}, a local structural-supervision baseline. Their results are evaluated on the same fixed Topo-RS-Hard records. The clDice model is trained as a supervised topology-loss baseline and is not initialized from a preference-optimized checkpoint. The continued-SFT control reuses exactly the 8762 Topo-Prior samples and matches the number of updates and optimizer schedule, but replaces preference optimization with BCE and Dice supervision.

We further include a disagreement-masked BCE control that matches the spatial correction support of \topodpo{} while replacing reference-relative preference weighting with ordinary supervised correction. This provides a stringent test of the contribution from the dense preference rule itself. For each Topo-Prior pair, we define
the disagreement indicator as
$D^{ij}=\mathbb{I}[Y_w^{ij}\neq Y_l^{ij}]$.
Let $\ell_{\mathrm{BCE}}$ denote the per-pixel BCE-with-logits loss.
The masked correction objective is
\begin{equation}
\mathcal{L}_{\mathrm{mask}}
=
\frac{1}{HW}
\sum_{i,j}
D^{ij}\,
\ell_{\mathrm{BCE}}
\!\left(M_\theta^{ij},Y_w^{ij}\right).
\label{eq:masked_bce}
\end{equation}
We deliberately retain the same $1/(HW)$ normalization as TopoDPO
rather than normalizing by the number of disagreement pixels,
$\sum_{i,j}D^{ij}$. The latter would additionally reweight training
examples according to disagreement area, confounding localized
supervision with a change in sample weighting. Apart from the
correction objective, this control uses exactly the same Topo-Prior
pairs, policy initialization, LoRA configuration, optimizer,
learning-rate schedule, number of updates, and supervised anchor as
TopoDPO.

A direct use of \eqref{eq:masked_bce}, however, gives a substantially different correction-gradient scale from dense DPO. We therefore match the two correction terms at the common initialization $\theta=\theta_{\mathrm{ref}}$ without tuning on validation data. At this initialization, policy and reference margins coincide, and direct differentiation of \eqref{eq:dense_dpo} yields an absolute logit-gradient magnitude of $\beta/(2HW)$ at every disagreement pixel. For masked BCE, the corresponding magnitude before scaling is $|\sigma(M_{\mathrm{ref}}^{ij})-Y_w^{ij}|/(HW)$. We compute once over the 8762 training pairs
\begin{equation}
\begin{aligned}
\bar g_{\mathrm{BCE}}
&=\frac{\sum_{n,i,j}D_n^{ij}\left|\sigma(M_{\mathrm{ref},n}^{ij})-Y_{w,n}^{ij}\right|}
{\sum_{n,i,j}D_n^{ij}},\\
\alpha_{\mathrm{mask}}
&=\frac{\beta}{2\bar g_{\mathrm{BCE}}},
\end{aligned}
\label{eq:mask_scale}
\end{equation}
and then keep $\alpha_{\mathrm{mask}}$ fixed for the complete run. The resulting control objective is
\begin{equation}
\begin{aligned}
\mathcal{L}_{\mathrm{ctrl}}
&=\alpha_{\mathrm{mask}}\mathcal{L}_{\mathrm{mask}}\\
&\quad+\lambda_{\mathrm{sft}}\left(\mathcal{L}_{\mathrm{BCE}}+\lambda_{\mathrm{dice}}\mathcal{L}_{\mathrm{Dice}}\right).
\end{aligned}
\label{eq:masked_control}
\end{equation}
Thus, at initialization, the two correction terms have the same average absolute logit-gradient magnitude on the same disagreement support. The scale in \eqref{eq:mask_scale} is computed only from training-pair logits and labels and is fixed for the complete run. This matched control compares reference-relative dense correction with conventional masked BCE under the same data, correction support, and initial average gradient scale.

\begin{table*}[!t]
\centering
\caption{Results on Topo-RS-Hard (4590 records). Topo-Prior supervised controls use the same 8762 samples and matched training schedules; masked BCE also matches the initial correction-gradient scale. Best values are bold.}
\label{tab:main_results}
\begingroup
\renewcommand{\arraystretch}{1.22}
\begin{tabular*}{0.99\textwidth}{@{\extracolsep{\fill}}lcccccc@{}}
\toprule
Method & Alignment/loss & IoU & F1 & Precision & $\Delta\beta_0$ & Corner recall\\
\midrule
SFT baseline (2025)~\cite{sa2va} & BCE+Dice & 47.88 & 56.04 & 59.97 & 2.77 & 63.59\\
SFT + clDice (2021)~\cite{cldice} & Topology loss & 48.57 & 56.76 & \textbf{60.54} & 2.37 & 64.51\\
SFT + SCNP (2026)~\cite{scnp2026} & Local structural loss & 48.90 & 57.12 & 59.95 & 2.27 & 65.06\\
Topo-Prior continued SFT (ours) & BCE+Dice & 49.63 & 57.92 & 60.08 & 2.12 & 65.38\\
Topo-Prior masked BCE (ours) & Disagreement pixels & 50.72 & 59.03 & 60.18 & 1.18 & 67.21\\
\midrule
Vanilla DPO (2023)~\cite{dpo} & Image level & 50.91 & 59.16 & 59.62 & 1.34 & 67.58\\
RN-DPO (2026)~\cite{rndpo2026} & Region-normalized DPO & 49.12 & 57.32 & 60.31 & 2.08 & 65.19\\
\method{} (Ours) & Pixel level & \textbf{51.24} & \textbf{59.45} & 60.27 & \textbf{0.90} & \textbf{67.83}\\
\bottomrule
\end{tabular*}
\endgroup
\end{table*}

\begin{table*}[!t]
\centering
\caption{Paired image-cluster bootstrap comparisons with preference-optimization baselines on Topo-RS-Hard. We resample 2581 unique images with replacement for 50\,000 replicates, keeping multiple referring expressions from one image in the same cluster. Differences are \method{} minus the indicated baseline; positive values favor \method{} for IoU, F1, precision, and corner recall, while negative values favor \method{} for Betti-0 error. Differences and intervals for percentage metrics are expressed in percentage points; Betti-0 error is an absolute difference.}
\label{tab:bootstrap}
\begingroup
\renewcommand{\arraystretch}{1.14}
\begin{tabular*}{0.90\textwidth}{@{\extracolsep{\fill}}llcc@{}}
\toprule
Baseline & Metric & Mean difference & 95\% confidence interval\\
\midrule
\multirow{5}{*}{Vanilla DPO (2023)~\cite{dpo}}
& IoU & $+0.33$ & $[+0.24,+0.43]$\\
& F1 & $+0.30$ & $[+0.20,+0.40]$\\
& Precision & $+0.65$ & $[+0.52,+0.79]$\\
& $\Delta\beta_0$ & $-0.44$ & $[-0.50,-0.38]$\\
& Corner recall & $+0.26$ & $[+0.04,+0.48]$\\
\midrule
\multirow{5}{*}{RN-DPO (2026)~\cite{rndpo2026}}
& IoU & $+2.12$ & $[+1.86,+2.37]$\\
& F1 & $+2.13$ & $[+1.86,+2.41]$\\
& Precision & $-0.04$ & $[-0.29,+0.24]$\\
& $\Delta\beta_0$ & $-1.18$ & $[-1.34,-1.03]$\\
& Corner recall & $+2.64$ & $[+2.22,+3.07]$\\
\bottomrule
\end{tabular*}
\endgroup
\end{table*}

\subsubsection{Metrics}
We report per-instance IoU, F1, and precision together with Betti-0 error and macro-corner recall. IoU measures intersection divided by union, while F1 is the harmonic mean of pixel precision and recall. Betti-0 error is the absolute difference in 8-connected component counts, and macro-corner recall follows \eqref{eq:corner}. Binary predictions use a 0.5 threshold, and the main table reports the arithmetic mean over the fixed records. The metric set jointly evaluates semantic coverage, component-count fidelity, and recovery of coarse annotated geometry.

\subsection{Results on the Structural Hard Set}\label{sec:hard_results}

Table~\ref{tab:main_results} compares the proposed method with supervised, topology-loss, and preference-optimization baselines, including the recent RN-DPO~\cite{rndpo2026} and SCNP~\cite{scnp2026}, on the fixed Topo-RS-Hard diagnostic set.

Table~\ref{tab:bootstrap} uses paired image-cluster resampling so that repeated prompts from the same aerial scene remain in the same bootstrap unit. Against Vanilla DPO~\cite{dpo}, the 95\% confidence intervals exclude zero for all five metrics. Against RN-DPO~\cite{rndpo2026}, the intervals exclude zero for IoU, F1, Betti-0 error, and corner recall, whereas the precision interval $[-0.29,+0.24]$ includes zero. Using the two-decimal values in Table~\ref{tab:main_results}, the precision mean difference is $-0.04$ percentage points.

The matched supervised controls in Table~\ref{tab:main_results} progressively isolate the contribution of the proposed alignment rule. Continued SFT reuses the same Topo-Prior samples and optimization schedule with ordinary BCE+Dice supervision. The disagreement-masked BCE control further matches the exact correction support, $1/(HW)$ normalization, supervised anchor, and initial average absolute logit-gradient scale of \topodpo{}. Under these matched conditions, \method{} improves IoU from 50.72\% to 51.24\% and reduces Betti-0 error from 1.18 to 0.90 relative to masked BCE. This matched mean-performance comparison, together with the factorial ablation in Table~\ref{tab:ablations}, helps distinguish dense preference weighting from localized error-pixel supervision.

Table~\ref{tab:main_results} shows higher IoU and lower Betti-0 error than the evaluated structural-loss baselines and RN-DPO~\cite{rndpo2026}, although clDice~\cite{cldice} has the highest precision. Relative to image-level DPO~\cite{dpo}, \method{} improves IoU from 50.91\% to 51.24\% and Betti-0 error from 1.34 to 0.90, with a smaller corner-recall gain (67.58\% to 67.83\%). Comparisons with other published baselines are method-level rather than fully matched.

\begin{figure*}[!t]
\centering
\begin{minipage}{0.42\textwidth}
    \centering
    \includegraphics[width=\linewidth]{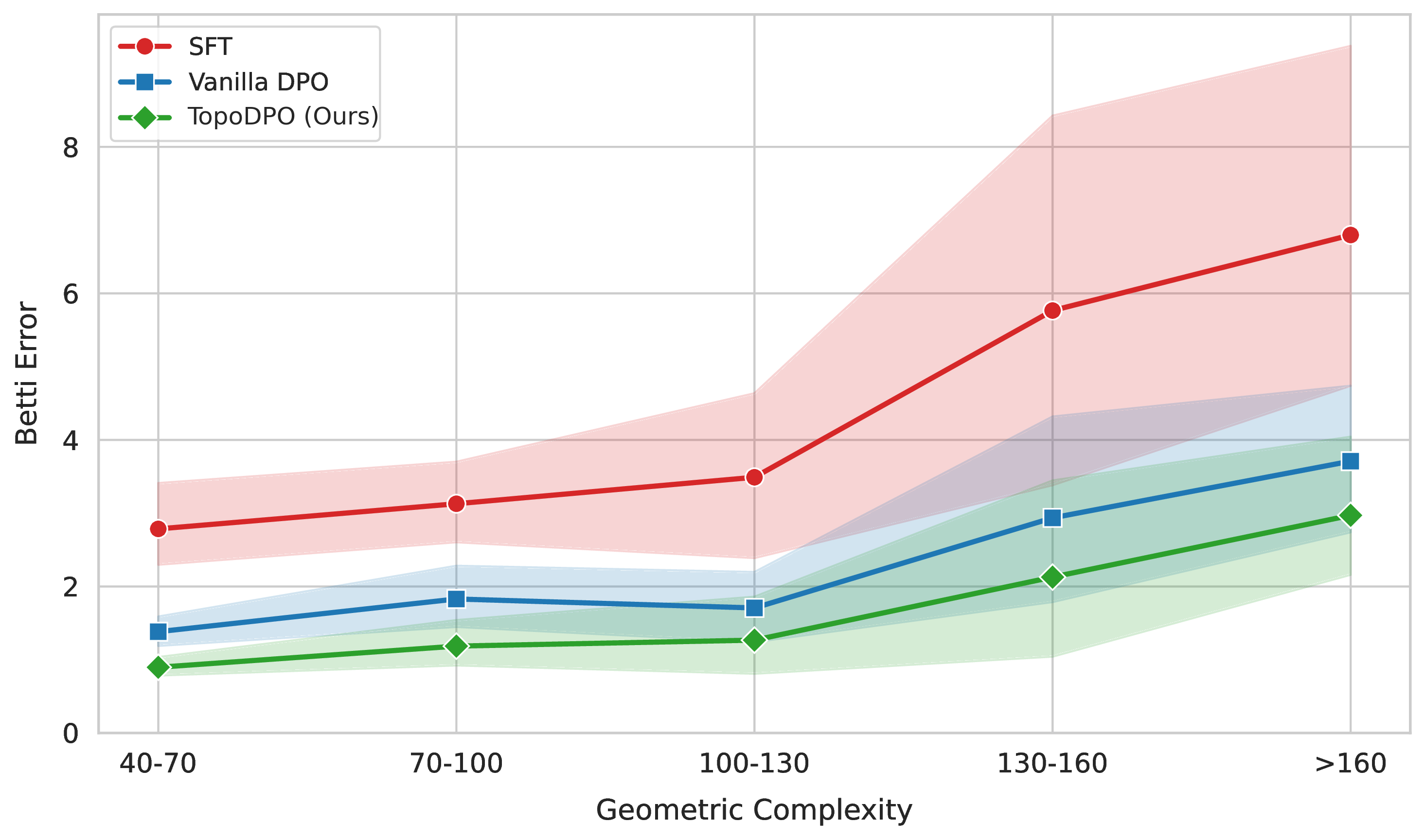}\\[2pt]
    {\footnotesize\textbf{(a)}}
\end{minipage}\hfill
\begin{minipage}{0.54\textwidth}
    \centering
    \includegraphics[width=\linewidth]{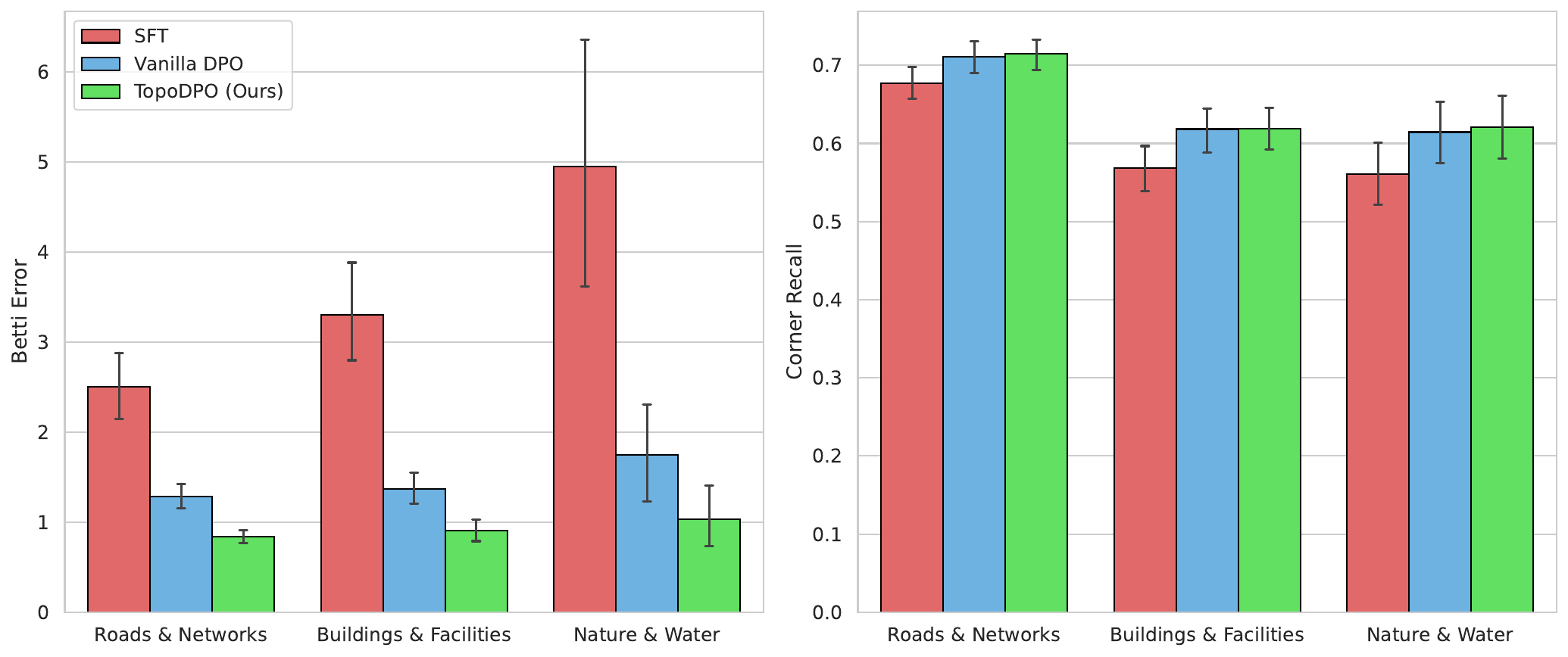}\\[2pt]
    \makebox[0.5\linewidth][c]{{\footnotesize\textbf{(b)}}}%
    \makebox[0.5\linewidth][c]{{\footnotesize\textbf{(c)}}}
\end{minipage}
\caption{Stratified Topo-RS-Hard analysis:
(a) Betti-0 error by geometric complexity;
(b) Betti-0 error by target category;
(c) corner recall by target category.
Shaded regions and error bars indicate 95\% confidence intervals.}
\label{fig:combined_analysis}
\end{figure*}

Fig.~\ref{fig:combined_analysis} stratifies Topo-RS-Hard by geometric complexity and coarse target category. The complexity statistic $P^2/A$ increases for elongated objects, highly indented boundaries, and shapes with large perimeter relative to area. Betti-0 error rises with this statistic for every model, especially in the highest-complexity bins, while \method{} remains below the SFT and image-level DPO curves across all displayed bins. The consistent separation across complexity levels shows that the aggregate connectivity gain is distributed across a broad range of difficult object geometries.

Across roads and networks, buildings and facilities, and nature and water, \method{} has the lowest Betti-0 error in each displayed group, suggesting that the connectivity gain is not confined to a single target family. Corner recall complements this connectivity analysis for artificial facilities.

\subsubsection{Unfiltered Full-Validation-Pool Stratification}

\begin{figure}[!t]
\centering
\includegraphics[width=\columnwidth]{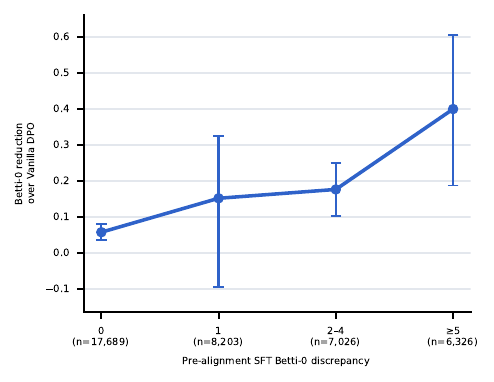}
\caption{Unfiltered full-validation-pool stratification by pre-alignment structural-error severity. All 39\,244 Aerial-D and RISORS validation records are grouped by the frozen SFT model's Betti-0 discrepancy, with no IoU-based inclusion filter. The vertical axis reports the reduction in Betti-0 error of \method{} relative to Vanilla DPO; positive values favor \method{}. Error bars are 95\% image-cluster bootstrap confidence intervals.}
\label{fig:fullpool_stratification}
\end{figure}

We expand the analysis to all 39\,244 Aerial-D+RISORS validation records without the Topo-RS-Hard or IoU golden-zone filters, grouping records only by the frozen SFT model's pre-alignment Betti-0 discrepancy. Because grouping is fixed before preference alignment and uses neither Vanilla DPO nor \method{} outputs, it provides an unfiltered view of how the relative structural effect varies with baseline error.

Fig.~\ref{fig:fullpool_stratification} shows that the Betti-0 advantage of \method{} becomes larger as the pre-alignment structural error increases. When the SFT Betti-0 discrepancy is 0, \method{} reduces Betti-0 error by 0.058 on average relative to Vanilla DPO (95\% image-cluster bootstrap confidence interval $[0.035,0.080]$). The reduction is 0.152 for discrepancy 1 ($[-0.095,0.326]$), 0.176 for discrepancy 2--4 ($[0.103,0.250]$), and 0.400 for discrepancy $\geq5$ ($[0.187,0.605]$). Thus, the same severity-dependent structural correction observed on Topo-RS-Hard remains visible over the unfiltered full validation population, with the clearest separation occurring on moderate and severe component-count failures.

\subsection{Ablation and Parameter Sensitivity}\label{sec:ablation}

\begin{table*}[!t]
\centering
\caption{Factorial ablations and anchor sensitivity on Topo-RS-Hard}
\label{tab:ablations}
\begingroup
\renewcommand{\arraystretch}{1.22}
\begin{tabular*}{0.96\textwidth}{@{\extracolsep{\fill}}lllcccc@{}}
\toprule
ID & Pair selection & Objective/configuration & IoU & F1 & $\Delta\beta_0$ & Corner recall\\
\midrule
A1 & Random & Image-level DPO & 49.18 & 58.31 & 2.06 & 64.72\\
A2 & Random & Pixel-level DPO & 49.52 & 58.78 & 1.85 & 65.12\\
A3 & Topo-Prior & Image-level DPO & 50.91 & 59.16 & 1.34 & 67.58\\
A4 & Topo-Prior & Pixel-level DPO & \textbf{51.24} & \textbf{59.45} & \textbf{0.90} & \textbf{67.83}\\
\midrule
B1 & Topo-Prior & $\lambda_{\mathrm{sft}}=0$ & 48.93 & 58.85 & 1.55 & 65.87\\
B2 & Topo-Prior & $\lambda_{\mathrm{sft}}=0.1$ & \textbf{51.24} & \textbf{59.45} & \textbf{0.90} & \textbf{67.83}\\
B3 & Topo-Prior & $\lambda_{\mathrm{sft}}=1.0$ & 50.85 & 58.90 & 1.02 & 67.75\\
\bottomrule
\end{tabular*}
\endgroup
\end{table*}

Table~\ref{tab:ablations} crosses pair selection with objective granularity on Topo-RS-Hard. Pixel-level DPO improves Betti-0 error under both pair-selection rules, while Topo-Prior improves both image-level and pixel-level objectives. The supervised-anchor study further shows that balancing preference alignment with segmentation supervision is important for retaining area accuracy while improving structure.

\subsubsection{Factorial Effects}
With Topo-Prior fixed, moving from image-level to pixel-level penalization decreases Betti-0 error from 1.34 to 0.90 and increases IoU from 50.91\% to 51.24\%. Under random selection, pixel-level penalization gives a smaller change, reducing Betti-0 error from 2.06 to 1.85 and increasing IoU from 49.18\% to 49.52\%. Conversely, Topo-Prior improves either objective relative to random pairs. These crossed effects support complementary roles: topology-aware selection determines which failures receive supervision, while dense DPO determines how local margins within each pair are weighted.

\subsubsection{Supervised-Anchor Weight}
Block B tests $\lambda_{\mathrm{sft}}\in\{0,0.1,1.0\}$. Removing the anchor lowers IoU and increases Betti-0 error, showing the importance of retaining supervised segmentation guidance during preference alignment. Increasing the weight to 1.0 constrains adaptation and yields slightly weaker component fidelity. Among the tested settings, $\lambda_{\mathrm{sft}}=0.1$ provides the best balance, reaching 51.24\% IoU and 0.90 Betti-0 error.

\begin{figure*}[!t]
\centering
\includegraphics[width=0.9\textwidth]{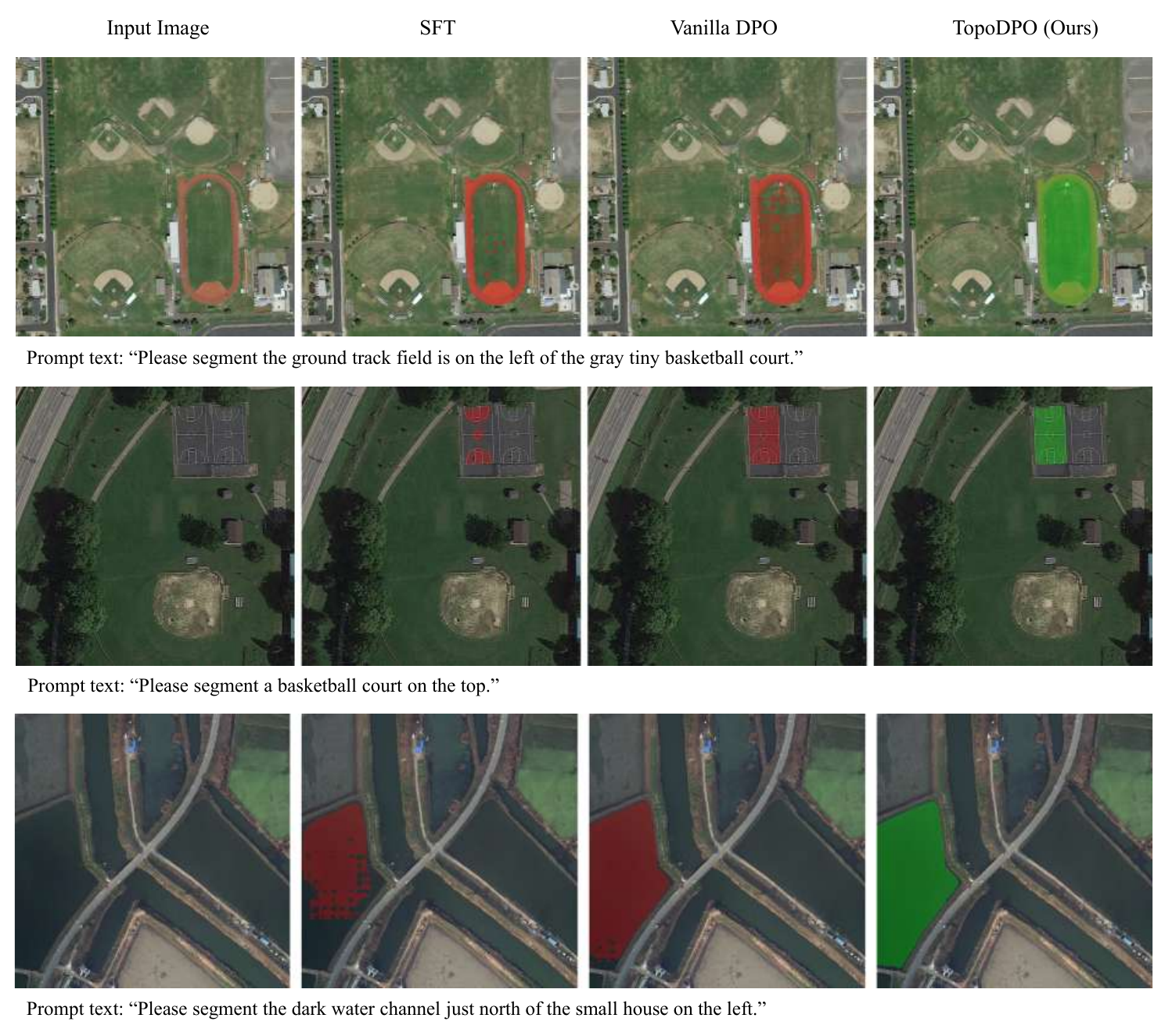}
\caption{Qualitative comparison on structurally complex examples. Red baseline masks contain fragmentation, holes, or boundary smoothing; green \method{} masks are more coherent in the displayed cases.}
\label{fig:qualitative_comparison}
\end{figure*}

\subsection{Qualitative Results}
Fig.~\ref{fig:qualitative_comparison} shows representative non-compact targets. The baselines broadly localize the referred objects but leave fragmentation, extent errors, holes, or rounded artificial corners, while \method{} produces more coherent masks. In the ground-track-field example, SFT and Vanilla DPO preserve the correct instance but contain interior omissions; for the basketball court, image-level DPO extends into neighboring structure; and for the narrow water channel, small missing sections alter continuity. These examples qualitatively match the measured improvements in connectivity and coarse geometry.

\subsection{General Multimodal Capability Retention}
To examine retention of general remote-sensing multimodal capability after segmentation-specific preference alignment, we evaluate the models on a fixed random sample of 1000 held-out EarthVQA validation questions~\cite{earthvqa}. EarthVQA supervision is included during the preceding SFT stage but is not used during preference alignment. SFT obtains 71.90\% accuracy and \method{} obtains 72.30\%. The comparable held-out performance indicates that the topology-sensitive alignment stage preserves the general multimodal capability learned during SFT.

\subsection{Interpretation and Scope}
The two preference objectives have identical local-margin gradients at their shared initialization. In particular, when the policy equals the frozen reference, every local margin and its spatial mean are zero, so both (13) and (14) assign a weight of $\beta/(2HW)$ to each margin. Their difference arises only after optimization creates spatially heterogeneous policy--reference margins. Accordingly, the gradient analysis characterizes how subsequent updates are weighted; it does not by itself establish that structural pixels received larger weights throughout training. The matched image-level comparison and the factorial ablation instead evaluate the net effect of this change in weighting under the same preference pairs and training protocol.

The evaluation measures different, noninterchangeable properties of the predicted masks. Betti-0 error compares component counts, so equal values do not imply the same component identities or connections. Macro-corner recall tests recovery of simplified annotated corners but does not penalize additional predicted corners directly. Consequently, the reported structural results should be interpreted as improved component-count fidelity and coarse geometric recall, not as a guarantee of topological equivalence or route connectivity. Moreover, Topo-RS-Hard intentionally selects non-compact validation targets with structural errors diagnosed before alignment. It provides a stress test of the intended failure mode rather than an estimate of the prevalence of these errors in arbitrary overhead images. The full-pool stratification offers a complementary check for component-count improvements, while broader claims about segmentation quality outside the evaluated datasets would require additional evidence.

\section{Conclusion}
We presented \method, a topology-aware dense preference-alignment framework for aerial referring segmentation. Topo-Prior mines structurally informative mask preferences offline, while \topodpo{} preserves local preference margins through nonlinear weighting before spatial aggregation. On Topo-RS-Hard, \method{} reduces Betti-0 error by 32.8\% relative to image-level DPO while improving IoU from 50.91\% to 51.24\%; paired image-cluster intervals against image-level DPO exclude zero for all five reported differences. Matched disagreement-masked BCE controls and factorial ablations further separate localized supervision, topology-aware pair selection, and dense preference weighting. Across all 39\,244 validation records, the clearest structural gains occur on samples with larger pre-alignment component-count errors.

\bibliographystyle{IEEEtran}
\bibliography{references}
\end{document}